\documentclass[10pt,twocolumn]{article}

\usepackage[letterpaper,margin=0.72in,columnsep=0.24in]{geometry}

\usepackage{booktabs}
\usepackage{multirow}
\usepackage{graphicx}

\usepackage{amsmath}
\usepackage{amssymb}

\usepackage{microtype}
\usepackage{xcolor}

\usepackage{url}
\usepackage{hyperref}

\usepackage{enumitem}

\usepackage{caption}
\usepackage{subcaption}

\usepackage[authoryear,round]{natbib}

\usepackage{placeins}
\usepackage{tabularx}
\usepackage{array}

\hypersetup{
    colorlinks=true,
    linkcolor=black,
    citecolor=black,
    urlcolor=black,
    pdftitle={Governance-State Corruption in Long-Term Memory for LLM Agents: A Cross-Model Stress Test},
    pdfauthor={Bowen Liu, Youting Wang, Bonan Shen, Dingyan Shang}
}

\title{%
\bfseries
Governance-State Corruption in Long-Term Memory\\
for LLM Agents: A Cross-Model Stress Test
}

\author{%
Bowen Liu$^{a,*}$ \quad Youting Wang$^{b}$ \quad Bonan Shen$^{c}$ \quad Dingyan Shang$^{d}$\\[3pt]
{\small $^{a}$Independent Researcher, South San Francisco, CA, USA}\\[-1pt]
{\small $^{b}$Independent Researcher, Mountain View, CA, USA}\\[-1pt]
{\small $^{c}$Independent Researcher, Long Island City, NY, USA}\\[-1pt]
{\small $^{d}$Independent Researcher, Frisco, TX, USA}\\[3pt]
{\small $^{*}$Corresponding author: Bowen Liu (\href{mailto:chris3494285@gmail.com}{chris3494285@gmail.com})}%
}

\date{}

\begin{document}

\maketitle

\begin{abstract}
An LLM agent can retrieve a memory correctly yet use it incorrectly after
its validity state changes. We study \emph{governance-state corruption}:
faults that alter which retained records enter model-visible context while
stored content and tasks remain fixed. Across six providers, 731
benchmark-grounded bases yield 2,080 matched fault pairs. In offline
control-plane checks, corruption changes the operative projection in
96.25\% of pairs, while oracle-backed enforcement restores it in 100\%.
For downstream evaluation, a deterministically selected, non-probability
LongMemEval/Memora panel resolves 298 of 300 matched triplets (238 distinct
bases). Clean-minus-corrupted contrasts are $40.54$ percentage points in
LongMemEval binary correctness (95\% base-cluster interval $[30.32,50.71]$)
and $30.60$ points in rescaled Memora FAMA ($[23.31,37.69]$); both sources
also show recovery. OAKS Novel native option parity holds for all 6,714
resolved outputs. Full-cohort frozen-scorer corruption/recovery effects
($-27.55/+27.51$ points) remain descriptive proxies, not validated
native semantic population effects. These bounded findings motivate
evaluating memory control-state integrity alongside retrieval accuracy.
\end{abstract}

{\small
\par}

\section{Introduction}

Consider an assistant asked to list a user's current weekly to-do items.
In an audited \emph{Memora} case, two relevant items, $u$ and $v$, remain
stored. With source-supported governance metadata, both enter model-visible
context. When a missing-metadata fault excludes $u$, however, only $v$
remains visible to the model, although the task and stored item content do
not change. Oracle-backed restoration makes both items visible again. In
existing six-provider runs, frozen-proxy correctness follows a
correct--incorrect--correct pattern. This case illustrates a control-plane
failure; it does not establish native FAMA correctness independently
(Section~\ref{sec:m3-case-trace}).

We separate stored evidence (the \emph{data plane}) from lifecycle and
validity state (the \emph{control plane}) governing what enters the model's
context. Deletion, supersession, and temporal updates can invalidate retained
evidence without altering its content. Our question is: \emph{How does an
agent behave when retrieved memory content is held fixed but the state
controlling its use is corrupted?} Here, governance denotes
benchmark-supported currentness and lifecycle validity; it does not always
represent an explicit user authorization policy.

We create matched \emph{clean}, \emph{corrupted}, and \emph{enforced}
conditions for the same task, oracle, evidence, prompt, and model. Clean uses
source-supported control state; corrupted injects a supported fault; enforced
restores the source-supported state before context construction. Thus,
enforcement tests an \emph{oracle-backed upper bound}, not autonomous repair.
We study the pathway $G\!\rightarrow\!A\!\rightarrow\!C\!\rightarrow\!Y$
(governance, admission, model-visible context, output) while measuring
operative context directly rather than identifying each mediator separately.

Across LongMemEval~\citep{wu2025longmemeval},
Memora~\citep{uddin2026memora}, and OAKS~\citep{kim2026oaks},
731 benchmark-grounded bases yield 2,080 supported fault pairs and 4,891
conditions, evaluated with six providers. An offline audit finds projection
changes in 96.25\% of pairs and restoration in 100\%. A separate,
non-probability \emph{native-style} Formal-300 validation resolves 298
LongMemEval/Memora triplets (238 distinct bases): corruption reduces
LongMemEval binary correctness by 40.54 points and Memora FAMA by 30.60
points, with recovery under both metrics. The full-panel $-27.55/+27.51$
point effects are \emph{frozen deterministic-scorer proxies}, not validated
population-native effects; OAKS has a separate resolved-option parity check.

Our research questions examine (RQ1) control-plane and downstream sensitivity,
(RQ2) supported fault differences, (RQ3) event-associated differences, and
(RQ4) oracle-backed recovery and provider/source robustness. We contribute:
(i) a governance-state corruption formulation distinct from retrieval or
content errors; (ii) a matched sibling-intervention benchmark with explicit
source--event--fault support; (iii) six-provider frozen-proxy results and
source-specific native-style validation with dependence-aware resampling; and
(iv) audited context propagation and bounded recovery evidence. We explicitly
separate source-confounded descriptive contrasts from supported intervention
claims.

\section{Related Work}

\paragraph{Long-term agent memory.}
MemGPT introduced virtual-context management for persistent agents
\citep{packer2023memgpt}; Mem0, A-Mem, and LangMem develop extraction,
organization, consolidation, and retrieval mechanisms
\citep{chhikara2025mem0,xu2025amem,langchain2025langmem}.
LongMemEval and its successor evaluate retention, multi-session reasoning,
updates, and abstention~\citep{wu2025longmemeval,wu2026longmemevalv2}.
These systems and benchmarks mainly ask whether an agent \emph{retains or
retrieves} the right evidence. That is necessary but insufficient: a
retrieved item may be authentic evidence of an earlier state while no longer
being valid as the current answer. We evaluate this distinction without
modifying the content retrieved from the frozen benchmark.

\paragraph{Governance, forgetting, and temporal validity.}
GateMem studies shared-memory access control and active forgetting;
Memora evaluates deletion and update-aware memory through forgetting-sensitive
criteria~\citep{ren2026gatemem,uddin2026memora}.
Safety-governed memory and MemGuard propose verification, lifecycle signals,
and controlled admission~\citep{lam2026ssgm,wang2026memguard}.
MemConflict distinguishes temporal validity, factual correctness, and
contextual applicability~\citep{tao2026memconflict}. These works motivate
explicit control state but differ from our \emph{paired state-corruption}
stress test, which fixes task and stored evidence and changes the state
selecting what becomes operative. In particular, Memora's FAMA evaluates
memory-presence and forgetting-absence criteria, while our full-panel
frozen scorer is a Boolean proxy. This makes scoring correspondence part of
the empirical question rather than an assumption. Our independent
native-style subset is explicitly limited to cases where the corresponding
source-specific evaluation can be carried out.

\paragraph{Control-plane foundations.}
Complete mediation and fail-safe defaults require trustworthy authorization
checks~\citep{saltzer1975}; Usage Control (UCON) treats access-related
attributes and decisions as mutable~\citep{park2004}. We adapt this
control-state perspective to agent memory as a
$G\rightarrow A\rightarrow C\rightarrow Y$ pipeline.
This lineage motivates a measurable boundary between lifecycle state,
admission, and the resulting context, rather than assuming the language
model can repair invalid evidence after receiving it. Our work neither
introduces a new access-control model nor demonstrates autonomous recovery:
it measures projection integrity, scorer-dependent behavioral degradation,
and restoration when benchmark-supported control state is provided. Unlike
an end-to-end compliance benchmark, it deliberately leaves the original
memory evidence and task fixed, making the control-plane transformation
inspectable; it does not make event types or benchmark datasets causally
interchangeable.

\section{Problem Formulation}

We distinguish retained memory evidence from the governance state that
determines whether it can enter model-visible context.

\subsection{Memory Content and Governance State}

Let a benchmark base example be
\[
b = (q, X, y^\star),
\]
where $q$ is the downstream query or task, $X$ denotes the benchmark-native
memory evidence associated with that task, and $y^\star$ is the
benchmark-supported scoring target.

We represent an individual memory item as
\[
m = (x,g),
\]
where $x$ is the underlying content and $g$ is governance state associated
with its use.
Governance state is source-dependent: it may encode whether a memory is
current, active, superseded, deleted, or otherwise admissible for the present
task.

For a collection of retrieved memory items $R \subseteq X$, let
\[
G = \{g_m : m \in R\}
\]
denote the corresponding governance state.

A retrieved item can therefore be factually accurate yet invalid to use.

\subsection{Governance-State Corruption}

Let $G^{\mathrm{clean}}$ denote the benchmark-supported governance state.
A fault operator $f$ transforms it into
\[
G^{\mathrm{corr}}_f
=
\phi_f\!\left(G^{\mathrm{clean}}\right),
\]
while leaving the underlying benchmark task $(q,X,y^\star)$ unchanged.

We study four recurring fault operators:

\begin{itemize}[leftmargin=*]
    \item \textbf{Missing:}
    governance information needed to determine admissibility or currentness is
    removed.

    \item \textbf{Stale:}
    a superseded or expired state is made operative.

    \item \textbf{Flipped:}
    a benchmark-supported governance decision or state is inverted.

    \item \textbf{Conflicting:}
    incompatible governance states are made simultaneously operative without
    preserving the benchmark-supported winner.
\end{itemize}

The concrete realization of each operator is source- and event-dependent:
a stale version in ordered narrative evidence is not an authorization
revocation, and a deletion state in Memora has different native semantics.
We instantiate interventions only when benchmark-native evidence identifies
both the clean state and its supported corrupted projection. No generic
policy ontology is imposed across the three benchmarks.

Importantly, $\phi_f$ is a control-state intervention rather than a content
corruption:
\[
X^{\mathrm{corr}} = X^{\mathrm{clean}} = X.
\]

\subsection{Four-Stage Control Pipeline}

We model downstream execution as
\[
G \rightarrow A \rightarrow C \rightarrow Y,
\]
where $G$ is governance state, $A$ is the enforcement or admission decision,
$C$ is model-visible context, and $Y$ is downstream behavior.

More explicitly,
\[
A = E(R,G,q),
\]
\[
C = H(R,A,q),
\]
and
\[
Y = M(q,C),
\]
where $E$ denotes enforcement, $H$ constructs the model-visible context, and
$M$ is the downstream language model.

We observe operative context changes, not every arrow as a separately
identified statistical mediator.

\begin{figure*}[t]
\centering
\includegraphics[width=\textwidth]{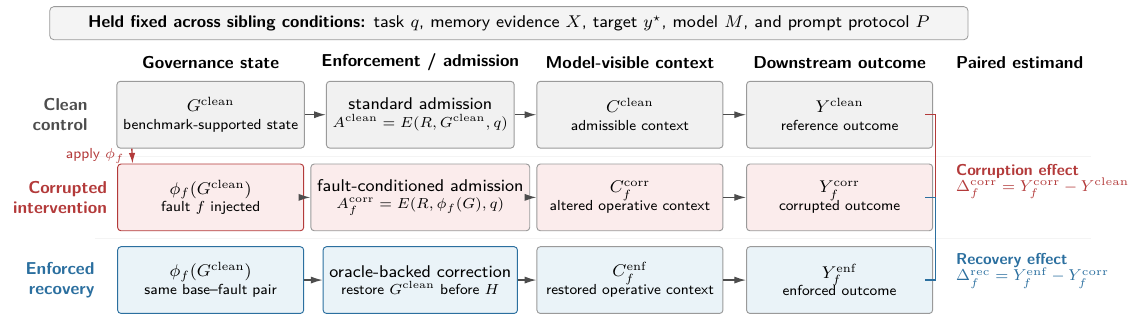}

\caption{
Control-plane intervention design.
The underlying benchmark task and memory evidence are held fixed.
The clean condition uses benchmark-supported governance state; the corrupted
condition injects a governance-state fault that can alter model-visible
context; and the enforced condition uses the same base--fault pair but restores
the benchmark-supported governance projection before context construction.
The enforced path is an oracle-backed mediation condition used to measure
recoverability, not an autonomous deployment-time repair algorithm.
}
\label{fig:control-plane}

\end{figure*}

Figure~\ref{fig:control-plane} shows the three siblings. Enforcement
measures recoverability when the correct control state is supplied.

\subsection{Controlled Sibling Intervention}

For each eligible base example, we construct clean, corrupted, and enforced
siblings.

For the enforced sibling, we define
\[
G^{\mathrm{enf}}_f
=
G^{\mathrm{clean}},
\]
where restoration is performed using the benchmark-supported governance state
identified during cohort construction.
The same base--fault pair and corruption specification are retained so that
recovery remains paired to the corresponding corruption intervention.

Across siblings, we hold fixed
\[
(q, X, y^\star, M, P),
\]
where $P$ denotes the frozen prompt protocol.

The paired contrasts are:
\[
Y^{\mathrm{corr}}_f - Y^{\mathrm{clean}}
\]
for corruption sensitivity, and
\[
Y^{\mathrm{enf}}_f - Y^{\mathrm{corr}}_f
\]
for oracle-backed recovery.

The exact correctness-based estimands are defined in Section~4.

\subsection{Scope of the Causal Claim}

\emph{Causal sensitivity} refers only to these fixed-input sibling
interventions. Oracle-backed enforcement does not show autonomous state
reconstruction, deployed fault prevalence, or downstream harm, and a fault
need not propagate to an output error.

\section{Method}

We instantiate the control-plane formulation as a frozen intervention benchmark
designed to isolate the effect of governance-state corruption from variation in
task content, oracle, prompt, or model identity.

\subsection{Benchmark Sources and Native Governance Evidence}

We construct the main cohort from LongMemEval, Memora, and OAKS, retaining only
examples whose source-native evidence identifies both a benchmark-supported
operative state and a corresponding governance intervention.

\paragraph{LongMemEval.}
LongMemEval~\citep{wu2025longmemeval} contributes long-horizon conversational
examples in which benchmark history distinguishes current from superseded
state, supporting temporal-state and update interventions.

\paragraph{Memora.}
Memora~\citep{uddin2026memora} contributes personalized memory examples with
source-native deletion, invalidation, preference, and update signals.

\paragraph{OAKS.}
OAKS (Online Adaptation to Continual Knowledge
Streams)~\citep{kim2026oaks} evaluates online adaptation under continually
evolving knowledge.
We use OAKS-Novel, constructed from 19 public-domain novels as ordered knowledge
streams whose answers may change with new evidence. It supports temporal-state
and update interventions; each base has one analysis event label. Across all
sources, questions and reference targets come from the source benchmark, not
synthetic question generation.

LongMemEval uses ordered conversational evidence of current versus
superseded facts; Memora supplies deletion and invalidation signals; OAKS
Novel supplies ordered narrative answer-state evidence. These operationalize
source-native \emph{validity}, not a shared consent or authorization ontology.

\subsection{Frozen Main Cohort}

All eligibility decisions and intervention manifests were frozen before the
formal cross-provider runs.

The 731-base manifest contains 380 deletion, 169 temporal, and 182 update
examples, yielding 4,891 conditions: 731 clean, 2,080 corrupted, and 2,080
enforced. 
All 380 deletion bases come from Memora, so deletion-versus-other-event
contrasts also compare different datasets and evaluation targets. Balanced
event sampling does not remove this source--event confounding.

Source-specific condition totals are 1,111 (LongMemEval), 2,660
(Memora), and 1,120 (OAKS); event totals are 2,660 deletion, 957 temporal,
and 1,274 update conditions. These are condition counts, not distinct bases.

\begin{table}[t]
\centering
\caption{Frozen source $\times$ event $\times$ fault support.
Fault counts denote pairs, not distinct bases; dashes are unsupported
operators. Event and source are not independently crossed.}
\label{tab:source-event-fault}
\footnotesize
\setlength{\tabcolsep}{2.7pt}
\begin{tabular}{llrrrrr}
\toprule
Source & Event & Bases & Miss. & Stale & Flip. & Conf. \\
\midrule
LongMemEval & Temporal & 113 & 113 & 113 & -- & -- \\
LongMemEval & Update & 78 & 78 & 78 & -- & 78 \\
Memora & Deletion & 380 & 380 & -- & 380 & 380 \\
OAKS Novel & Temporal & 56 & 56 & 56 & -- & 56 \\
OAKS Novel & Update & 104 & 104 & 104 & -- & 104 \\
\midrule
Total & & 731 & 731 & 351 & 380 & 618 \\
\bottomrule
\end{tabular}
\end{table}

Table~\ref{tab:source-event-fault} contains 14 populated
source--event--fault cells across five observed source--event strata; four
source--event combinations are absent and six further fault cells are
structurally inapplicable.
Exactly 618 bases support three faults and 113 LongMemEval temporal bases
two, giving $618\times3+113\times2=2{,}080$ pairs and
$731+2(2{,}080)=4{,}891$ conditions. Fault columns repeat bases; their
counts must not be interpreted as distinct examples. Deletion coincides
with Memora, while temporal and update have different source mixtures;
neither event balancing nor this table removes confounding.

The condition manifest is identical across all formal model runs;
provider execution never changes example eligibility, fault assignment,
event assignment, or scoring targets.

\subsection{Governance Fault Operators}

Missing, stale, flipped, and conflicting operators are deterministic
source-grounded state projections (Section~3). Across the corrupted and
enforced conditions, they account for 1,462 missing, 702 stale, 760 flipped,
and 1,236 conflicting rows. Source-native support is intentionally unequal;
no downstream outputs determine eligibility or fault assignment.

\subsection{Condition Construction and Oracle-Backed Enforcement}

We instantiate the clean, corrupted, and enforced siblings defined in
Section~3.4 from the frozen manifest.  Clean exposes the benchmark-supported
projection; corrupted applies a supported fault; enforced begins from the same
base--fault pair and restores the benchmark-supported projection before context
construction.

The enforcement transformation is deterministic and operator-matched.
A missing-state corruption is corrected using the supported governance fields
that were omitted;
a stale corruption restores the supported current state;
a flipped corruption restores the benchmark-supported decision or state; and
a conflicting corruption resolves the projection to the benchmark-supported
operative state.
We do not use the downstream answer produced by the model to perform this
correction.

No downstream answer is used for correction; model, task, evidence, oracle,
prompt, and inference settings are identical across siblings.

\subsection{Case Trace: Missing Governance Metadata}
\label{sec:m3-case-trace}

To make the intervention concrete, Table~\ref{tab:m3-case} gives a
content-abstracted trace of a previously selected \emph{Memora} weekly-todo
case in our offline paper-evidence audit. The fixed benchmark task asks for
current to-do items; we denote the two task items by $u$ and $v$ instead of
reprinting benchmark-derived text. The task, underlying evidence, reference
answer, and model configuration do not change across siblings. The missing
operator removes required governance metadata for $u$, changing the
operative projection from two items to one. Oracle-backed enforcement restores
that metadata and the two-item projection.

\begin{table*}[t]
\centering
\footnotesize
\setlength{\tabcolsep}{5pt}
\begin{tabularx}{\textwidth}{p{1.20in}XXX}
\toprule
\textbf{Pipeline stage} & \textbf{Clean} & \textbf{Corrupted: missing} & \textbf{Enforced} \\
\midrule
$G$: governance state & Required source-supported control fields present & Required metadata for item $u$ omitted & Benchmark-supported fields restored \\
$A$: admitted memory & Two items: $u,v$ & Only item $v$ & Two items: $u,v$ \\
$C$: model-visible memory & $[u,v]$ & $[v]$ & $[u,v]$ \\
$Y$: downstream frozen-proxy correctness & Six of six providers correct & Six of six providers incorrect & Six of six providers correct \\
\bottomrule
\end{tabularx}
\caption{De-identified audited Memora case trace (existing frozen outputs).
$u,v$ replace licensed to-do text; $Y$ is six-provider \emph{frozen-proxy}
correctness, not native FAMA. Per-case evidence is private; the audit is in
\texttt{scripts/audit\_paper\_evidence\_counts.py}.}
\label{tab:m3-case}
\end{table*}

The case-trace audit verifies the two--one--two context pattern. A separate,
read-only v2--v3 comparison confirmed identical case content, corruption
and restoration, and six-provider correct--incorrect--correct frozen-proxy
labels without new calls. Per-case source text and responses remain private;
the example does not establish native-style validity.

\subsection{Downstream Outcome and Paired Effects}

The full-panel downstream outcome is a frozen, source-aware
\emph{deterministic correctness proxy}, scored on generated responses; it is
not the benchmark-native semantic evaluator. In particular, the Memora proxy
checks extracted content facts as a Boolean decision, whereas native Memora
uses per-question memory-presence and forgetting-absence criteria (FAMA).
The full-panel proxy and native Memora score must not be interpreted as the
same measurement unit.

For base example $i$, fault $f$, and model $m$, let
\[
S^{\mathrm{clean}}_{i,m},
\quad
S^{\mathrm{corr}}_{i,f,m},
\quad
S^{\mathrm{enf}}_{i,f,m}
\in \{0,1\}
\]
denote the binary frozen-scorer decision in the three sibling conditions.

All full-panel contrasts below use the same frozen proxy, so pairing
controls input differences but does not by itself remove scorer bias.
We define
\[
\Delta^{\mathrm{corr}}_{i,f,m}
=
S^{\mathrm{corr}}_{i,f,m}
-
S^{\mathrm{clean}}_{i,m},
\]
and
\[
\Delta^{\mathrm{rec}}_{i,f,m}
=
S^{\mathrm{enf}}_{i,f,m}
-
S^{\mathrm{corr}}_{i,f,m}.
\]

A negative corruption contrast denotes a lower frozen proxy score under
corrupted state; a positive recovery contrast denotes a higher proxy score
under oracle-backed enforcement.

\subsection{Primary and Secondary Estimands}

Because event categories differ in size and not every base supports the
same number of faults, our primary proxy-score estimand gives \emph{equal
weight to each base within an event}, followed by equal weight to the three
events.  Let $\mathcal{B}_e$ be the set of bases assigned event $e$, and
$\mathcal{F}_i$ the supported fault operators for base $i$.  For provider
$m$, we first compute
\[
\bar{\Delta}^{\mathrm{corr}}_{i,m}
=\frac{1}{|\mathcal{F}_i|}
\sum_{f\in\mathcal{F}_i}
\Delta^{\mathrm{corr}}_{i,f,m},
\]
then the base-equal event mean
\[
\bar{\Delta}^{\mathrm{corr}}_{e,m}
=\frac{1}{|\mathcal{B}_e|}
\sum_{i\in\mathcal{B}_e}
\bar{\Delta}^{\mathrm{corr}}_{i,m},
\]
for $e\in\{\mathrm{deletion},\mathrm{temporal},\mathrm{update}\}$.
The reported event macro is
\[
\Delta^{\mathrm{corr}}_{\mathrm{macro},m}
=\frac{1}{3}\sum_e\bar{\Delta}^{\mathrm{corr}}_{e,m},
\]
with recovery defined by the identical three-stage aggregation.
The bootstrap resamples bases within events, preserving within-base fault
pairs, and then averages the three event means.

This differs from an unweighted \emph{fault-pair} event average, where bases
with three supported faults receive more weight than bases with two.
The distinction matters only for the temporal event: 113 LongMemEval
bases support two faults, while 56 OAKS Novel bases support three
(169 temporal bases but 394 fault pairs). Deletion and update each have
three supported faults per base, so the two event summaries coincide for
those events. Table~\ref{tab:event-breakdown} reports the \emph{base-equal}
event means from which the primary macro can be reconstructed. Original
fault-pair-weighted event summaries are retained in the aggregate audit
artifacts for transparency. Other source/fault descriptive breakdowns may
use their declared pair-level estimands rather than this event macro.
Providers are not treated as independent statistical replicates.

\subsection{Governance-Evidence Audit}
\label{sec:governance-audit-protocol}

The provider-independent audit
(\texttt{scripts/audit\_governance\_evidence.py}) checks frozen manifest
provenance and clean-state support: ordered current-state evidence for
LongMemEval, memory/forgetting fields for Memora, and document/chunk/answer
order for OAKS. It verifies task--oracle links, source locators,
operator-support rationales, and sibling structure without reading model
outputs. The 731/731 pass rate therefore measures \emph{deterministic
schema/provenance consistency}. Because the audit uses the same frozen
construction manifests, oracle states, and support declarations, it checks
traceability and internal consistency rather than independently establishing
the semantic correctness of those source interpretations. Licensed source
text is excluded from the aggregate-only release.

\subsection{Models and Inference Protocol}

We use DeepSeek V4 Pro, GPT-5.6 Sol, Claude Sonnet 5, Gemini 3 Flash,
Kimi K2.6 no-thinking, and Grok 4.3 no-reasoning, holding each provider's
inference configuration and source/task-class output budget fixed across
siblings. Condition labels are hidden; provider transport is normalized
before scoring.

\paragraph{Prompt and parameter provenance.}
All providers share \texttt{render\_prompt} in
\texttt{scripts/run\_final\_main\_smoke\_deepseek.py}: the model is told
to use only supplied context, avoid unsupported facts, and return a JSON
\texttt{answer} string. Requests include the task query and admitted context;
multiple-choice tasks provide labeled options. The public repo includes
rendering code, provider adapters, defaults, and token-budget configs,
not licensed instantiated prompts or complete per-request traces. For
example, Kimi no-thinking uses temperature 0.6/top-$p$ 0.95;
Grok sets \texttt{reasoning\_effort=none} without unsupported sampling
controls. API-side revisions, timestamps, and retry histories cannot be
independently reconstructed from shipped aggregates.

\subsection{Scoring and Failure Handling}

Full-panel scoring is deterministic, but its independent semantic validity
is source dependent. In a separate scorer-validity check of the frozen scorer against
separately generated native-style reference judgments, LongMemEval agrees on 60/96 \emph{answers}
(62.50\%, Cohen's $\kappa=0.259$). Memora's 508 reference units are \emph{atomic
criteria} spanning 42 answers, whereas the frozen scorer assigns one Boolean
per answer; no defensible same-unit agreement estimate exists for Memora.
The unaligned pooled 604-unit percentage is therefore not an interpretable
scorer-validity measure. We retain the original frozen results as
\emph{proxy-score effects}, without treating them as validated full-panel
semantic accuracy. OAKS has a separate native option-parity comparison,
described next.

After configured retries, genuine final provider or parse failures are retained
and scored as zero in the primary analysis.
A complete-case analysis is used as a sensitivity check.

Systematic truncation caused by insufficient output-token budget is treated as a
measurement artifact rather than model failure.
Affected rows are rerun under a documented larger budget and merged into the
formal artifact before analysis.

\subsection{Complementary Native-Style Validation and Sampling}
\label{sec:native-validation-method}

Formal-300 evaluates \emph{existing answers} with the frozen
\texttt{final-native-evaluation-v2} protocol. A seed-keyed greedy
stratification (seed 20261010) selects 25 complete answer triplets per
provider--source cell for six providers and two sources (LongMemEval,
Memora): 300 matched triplets. The full panel contains 12,480
provider--fault pairs, which have different counting units and do not
define a comparable sampling percentage. Selection is non-probabilistic;
the 300 triplets contain 239 distinct bases.
Two LongMemEval conflicting-fault blocks are unresolved, leaving 298
complete triplets, 238 bases, and 894 analyzed observations. Of 898
individually resolved observations, no incomplete triplet is used; one
recorded blinded human adjudication is reflected in the final version.

LongMemEval uses answer-level correctness/abstention judgments and
blinded semantic judging where necessary (\emph{native-style}, not the exact
upstream harness); Memora aggregates native atomic presence/forgetting
criteria into FAMA. These different outcome units are shown \emph{separately}.
Combined block-weighted point differences merely summarize two scaled
measures, not native accuracy or population effect. OAKS is absent from this
panel and instead has a deterministic native option check (6,714 resolved
and concordant of 6,720; six ambiguous/unscored).

\paragraph{Subgroup uncertainty.}
We preserve the 10,000-resample paired-block percentile bootstrap and
add a 10,000-resample (seed 20261010) base-cluster bootstrap within source,
retaining shared-provider blocks and sibling triplets. These intervals
describe stability \emph{within this selected non-probability panel}, not
population precision. Provider--source subsets of 24--25 blocks cannot
establish model rankings.

\subsection{Statistical Analysis}

Uncertainty is estimated using paired bootstrap resampling over base examples
with seed 42.
Sibling conditions from the same base remain grouped during resampling.
We report two-sided 95\% bootstrap confidence intervals for the primary
corruption and recovery effects.

We perform three robustness analyses.

\paragraph{Balanced event support.}
A balanced cohort contains 169 bases from each event category (507 total).

\paragraph{OAKS document clustering.}
Because multiple OAKS examples can originate from the same document, we repeat
the source-level analysis using document-level clustered resampling.

\paragraph{Leave-one-document-out analysis.}
We recompute the OAKS effect while omitting each source document in turn.

These robustness analyses are secondary to the frozen primary event-macro
estimand and are not used to redefine the main cohort after formal model
execution.

\subsection{Offline Control-Plane Propagation Diagnostic}

For the 2,080 supported base--fault pairs, we compare clean, corrupted,
and enforced \emph{ordered model-visible projections} $C^s_{i,f}$.
For base $i$ and fault $f$, we define projection-change and restoration
indicators
\[
Z^{\mathrm{chg}}_{i,f}
=
\mathbb{I}
\left[
C^{\mathrm{corr}}_{i,f}
\neq
C^{\mathrm{clean}}_{i,f}
\right],\qquad
Z^{\mathrm{rest}}_{i,f}
=
\mathbb{I}
\left[
C^{\mathrm{enf}}_{i,f}
=
C^{\mathrm{clean}}_{i,f}
\right].
\]

Projection-change and restoration rates are computed over the 2,080 unique
base--fault pairs rather than over provider-replicated rows.

We stratify the existing paired corruption contrast by projection change,
\[
\mathbb{E}
\left[
\Delta^{\mathrm{corr}}_{i,f,m}
\mid
Z^{\mathrm{chg}}_{i,f}=1
\right]
\quad\text{versus}\quad
\mathbb{E}
\left[
\Delta^{\mathrm{corr}}_{i,f,m}
\mid
Z^{\mathrm{chg}}_{i,f}=0
\right],
\]
and report induced-failure and recovery rates among clean-correct,
projection-changed cases.  These are descriptive provider-level diagnostics.
A separate admission-decision artifact was not persisted, so this analysis
tests the admitted/model-visible boundary rather than independently identifying
admission and context as separate mediators.

\subsection{Matched-Support Fault Analysis}

Because fault operators have different native support sets, we compare them on
bases supporting both interventions.

For provider $m$, base example $i$, and faults $f_1$ and $f_2$, we compute
\[
D_{i,m}(f_1,f_2)
=
\Delta^{\mathrm{corr}}_{i,f_1,m}
-
\Delta^{\mathrm{corr}}_{i,f_2,m}.
\]

Negative values indicate greater degradation under $f_1$.  The cohort contains
351 stale--missing and 238 stale--conflicting matched bases, but no
stale--flipped overlap.  We report provider-specific mean paired differences and
95\% paired-bootstrap intervals over matched base IDs (seed 42).

\subsection{Governance-Evidence Audit}

All 731 selected bases pass the frozen schema/provenance audit
(191 LongMemEval, 380 Memora, 160 OAKS), with zero schema-flagged
unsupported or provenance-missing bases. These \emph{rule-based}
consistency checks use the frozen construction evidence and support
declarations (Section~\ref{sec:governance-audit-protocol}).

\section{Results}

We first report the \emph{native-style validation panel}, then the
frozen deterministic-scorer panel. Both evaluate matched governance
interventions, but they differ in sampling, outcome units, and estimands.
Unless otherwise noted, RQ1--RQ4 and the remaining full-panel analyses use
\emph{frozen proxy-score} contrasts, not benchmark-native semantic accuracy.
Their bootstrap intervals describe sampling variation under that proxy,
not uncertainty caused by scoring mismatch.

\subsection{Native-Style Behavioral Validation on a Selected Panel}
\label{sec:formal-validation-results}

The fixed Formal-300 protocol selected 300 matched triplets from LongMemEval
and Memora across all six providers (25 blocks per provider--source cell).
The finalized native-style evaluation, including one blinded human
adjudication, resolves 298/300 triplets (238
distinct bases); two unresolved triplets are excluded. Source-specific results are
reported first in Table~\ref{tab:formal300-source-cluster}.
LongMemEval ($n=148$, binary correctness) has a $+40.54$-point
clean-minus-corrupted contrast and $+39.19$-point enforced-minus-corrupted
recovery. Memora ($n=150$, native continuous FAMA) has corresponding
$+30.60$-point damage and $+26.24$-point recovery. The 95\% base-cluster
resampling intervals for damage are $[30.32,50.71]$ and $[23.31,37.69]$
points, respectively; both sources show the same direction under their
\emph{own} native-style score units.

\begin{table*}[t]
\centering
\caption{Final Formal-300 native-style results on 298 resolved triplets.
Damage is clean minus corrupted; recovery is enforced minus corrupted,
scaled to points. LongMemEval is binary; Memora is continuous FAMA.
Brackets: descriptive 95\% base-cluster bootstrap intervals
(10,000 resamples; seed 20261010) on this selected panel.}
\label{tab:formal300-source-cluster}
\small
\setlength{\tabcolsep}{3pt}
\begin{tabular}{lrrrrrll}
\toprule
Source / unit & Triplets & Bases & Clean & Corrupted & Enforced & Damage [95\% interval] & Recovery [95\% interval] \\
\midrule
LongMemEval / binary & 148 & 105 & .709 & .304 & .696 & $+40.54$ $[30.32,50.71]$ & $+39.19$ $[28.85,49.66]$ \\
Memora / FAMA & 150 & 133 & .671 & .365 & .628 & $+30.60$ $[23.31,37.69]$ & $+26.24$ $[19.09,33.34]$ \\
\midrule
\shortstack[l]{\emph{Combined descriptive index}\\\emph{(mixed metrics)}} & 298 & 238 & .690 & .335 & .662 & $+35.54$ $[29.40,41.90]$ & $+32.67$ $[26.42,39.09]$ \\
\bottomrule
\end{tabular}
\end{table*}

The mixed-unit, equal-block-weight descriptive combined scores are
$0.6903$, $0.3349$, and $0.6617$ for clean, corrupted, and enforced.
Their clean-minus-corrupted difference is $+35.54$ points (original paired-block
95\% interval $[29.78,41.12]$; base-cluster interval $[29.40,41.90]$),
and the enforced-minus-corrupted difference is $+32.67$ points
(original $[26.94,38.36]$; base-cluster $[26.42,39.09]$).
These pooled values are a descriptive average of \emph{two different metrics},
not one benchmark-native accuracy. One observation required recorded human
adjudication; the older aggregate-only pre-adjudication analysis resolved
297 triplets and reported $+35.66$ points. The final 298-triplet analysis
uses 894 observations from complete triplets rather than all 898 individually
resolved observations.

Across the six generating-provider subsets, the native-style damage and
recovery contrasts are positive (Table~\ref{tab:formal300-provider-cluster}).
Each provider contributes only 49--50 triplets; base content is shared
across some providers, and source units differ. The base-cluster bootstrap
makes the pooled intervals modestly wider but does not reverse their
direction. Provider--source cells are particularly limited: DeepSeek--Memora
($+2.22$ damage, interval $[-10.73,15.12]$) and Grok--Memora
($+12.54$, $[-3.89,29.14]$) have intervals spanning zero. Accordingly,
these estimates do not support claims of provider superiority or uniformly
stable behavior in every provider--source cell.

\begin{table*}[t]
\centering
\caption{Selected Formal-300 provider-level damage and recovery.
Brackets: descriptive 95\% base-cluster intervals; each row averages
LongMemEval binary and Memora FAMA \emph{differences}, not a shared
native-accuracy metric. Cells are too small for provider rankings.}
\label{tab:formal300-provider-cluster}
\small
\setlength{\tabcolsep}{7pt}
\begin{tabular}{lrrll}
\toprule
Generating provider & Triplets & LME / Memora & Damage [95\% interval] & Recovery [95\% interval] \\
\midrule
GPT-5.6 Sol & 50 & 25 / 25 & $+51.13$ $[38.02,64.07]$ & $+50.03$ $[36.52,63.32]$ \\
Claude Sonnet 5 & 50 & 25 / 25 & $+50.60$ $[36.96,64.07]$ & $+40.38$ $[26.48,54.52]$ \\
DeepSeek V4 Pro & 49 & 24 / 25 & $+17.46$ $[6.25,29.23]$ & $+19.94$ $[8.87,31.68]$ \\
Gemini 3 Flash & 49 & 24 / 25 & $+44.04$ $[29.68,58.09]$ & $+41.52$ $[26.06,56.19]$ \\
Kimi K2.6 & 50 & 25 / 25 & $+31.53$ $[18.78,44.37]$ & $+26.11$ $[13.96,38.64]$ \\
Grok 4.3 & 50 & 25 / 25 & $+18.27$ $[3.78,32.55]$ & $+17.97$ $[5.34,31.05]$ \\
\bottomrule
\end{tabular}
\end{table*}

Recovery does not always regain clean performance. Memora's enforced-minus-clean
residual is $-4.35$ points (base-cluster interval $[-8.29,-0.66]$);
LongMemEval's is $-1.35$ points ($[-5.88,3.29]$). The separate native-style
panel is a deterministically stratified, non-probability selection with no
OAKS cases; its resampling intervals are descriptive of this panel, not
population confidence statements. It \emph{neither estimates nor calibrates}
the full-panel $-27.55$-point proxy effect, whose magnitude differs by
roughly eight points and uses another scorer and weighting scheme.

\subsection{Independent Scorer-Validity and OAKS Option Checks}

A separate scorer-validity audit compares frozen deterministic decisions
against native-style reference judgments. The \emph{same-unit} LongMemEval result
is 60/96 agreeing answer-level decisions (62.50\%, Cohen's $\kappa=0.259$),
with known paraphrase and abstention disagreements. For Memora, 508 atomic
criteria cover only 42 answers and cannot be compared as 508 independent
Boolean decisions; neither their raw agreement nor a 604-unit pooled rate
is valid answer-level calibration. The full-panel LongMemEval/Memora score
magnitudes remain proxy-dependent.

OAKS Novel uses a native deterministic option-comparison rule. On 6,720
OAKS response rows, 6,714 are resolved and \emph{all 6,714} agree with the
native comparator; the remaining six are ambiguous and excluded from the
agreement denominator. OAKS option-score parity is thus independently checked
for the resolved rows. This does not turn the combined full-panel
LME/Memora/OAKS event macro into a validated common semantic outcome.

\subsection{RQ1: Negative Corruption Effects under the Frozen Scorer}

Under the frozen proxy scorer, the corruption contrast is negative for every
provider in the six-provider panel. These are historical measured contrasts,
not six independently validated semantic-correctness estimates.

\begin{table*}[t]
\centering
\caption{
Six-provider frozen deterministic-scorer results (not independently
validated native semantic accuracy). Clean, corrupted, and enforced columns
report overall condition-level proxy-score rates (\%).
Corruption and recovery columns use base-equal fault averaging within each
event, followed by equal weighting of the three event means (Table~\ref{tab:event-breakdown});
values are percentage points; brackets denote 95\% paired-bootstrap
confidence intervals.
Because the accuracy columns and effect columns use different aggregation
schemes, their arithmetic differences need not equal the reported macro
effects.
}
\label{tab:main-results}
\small
\begin{tabular}{lrrrcc}
\toprule
Model & Clean & Corrupted & Enforced &
Corruption macro & Recovery macro \\
\midrule
DeepSeek V4 Pro
& 45.96 & 19.47 & 47.07
& $-28.05$ $[-31.30,-24.82]$
& $+28.10$ $[24.97,31.21]$ \\

GPT-5.6 Sol
& 43.23 & 17.02 & 43.51
& $-31.10$ $[-34.11,-28.09]$
& $+31.11$ $[28.15,34.07]$ \\

Claude Sonnet 5
& 43.23 & 18.03 & 43.85
& $-28.19$ $[-31.30,-25.02]$
& $+28.36$ $[25.51,31.23]$ \\

Gemini 3 Flash
& 43.78 & 16.35 & 44.13
& $-31.88$ $[-35.00,-28.75]$
& $+31.29$ $[28.21,34.36]$ \\

Kimi K2.6 no-thinking
& 42.95 & 19.23 & 45.29
& $-27.63$ $[-30.78,-24.46]$
& $+28.59$ $[25.75,31.42]$ \\

Grok 4.3 no-reasoning
& 28.59 & 12.02 & 28.56
& $-18.47$ $[-21.82,-15.14]$
& $+17.58$ $[14.65,20.54]$ \\
\bottomrule
\end{tabular}
\end{table*}

The primary event-macro corruption effect ranges from
$-31.88$ percentage points for Gemini 3 Flash to
$-18.47$ points for Grok 4.3.
The descriptive six-provider mean is
\[
-27.55\ \text{percentage points}.
\]

All six provider-level frozen-proxy intervals exclude zero.

The direction is consistent across providers for this fixed scoring rule.
The native-style panel measures a separate selected subset and does not
calibrate this six-provider table.

\subsection{RQ2: Fault Semantics Are Associated with Different Severities}

\begin{table*}[t]
\centering
\caption{
Observed frozen-scorer proxy effect by governance fault operator (points).
Fault operators have different native support sets, so these estimates describe
severity within the supported benchmark instances rather than a balanced causal
comparison among fault classes.
}
\label{tab:fault-breakdown}
\small
\begin{tabular}{lrrrr}
\toprule
Model & Conflicting & Flipped & Missing & Stale \\
\midrule
DeepSeek V4 Pro       & -5.50 & -37.37 & -33.11 & -45.30 \\
GPT-5.6 Sol           & -8.41 & -19.47 & -30.23 & -58.40 \\
Claude Sonnet 5       & -6.15 & -27.11 & -30.64 & -49.29 \\
Gemini 3 Flash        & -8.25 & -26.58 & -32.56 & -55.84 \\
Kimi K2.6 no-thinking & -2.59 & -26.58 & -30.10 & -50.71 \\
Grok 4.3 no-reasoning & -5.83 & -19.74 & -19.43 & -32.19 \\
\bottomrule
\end{tabular}
\end{table*}

Observed degradation varies substantially by fault semantics.
Among the benchmark-supported instances for each operator, stale corruption has
the largest negative effect for every provider, ranging from
$-32.19$ to $-58.40$ points.
Conflicting faults have the smallest observed degradation across the panel,
while flipped and missing faults occupy an intermediate regime.

Because the benchmark does not support every fault operator on the same set of
base examples, these values should not be interpreted as a randomized
head-to-head comparison among fault classes.
The stable ordering nevertheless shows that the evaluated stale-state
interventions are associated with particularly large downstream degradation
across all six providers.

To reduce the possibility that this pattern is explained by differences in
fault-support composition, we repeat the stale comparisons on matched bases.
Across the 351 bases supporting both stale and missing faults, stale produces
an additional 12.25--13.68 percentage-point degradation across the six
providers, with every provider-specific paired-bootstrap 95\% confidence
interval strictly below zero.

Across the 238 bases supporting both stale and conflicting faults, stale
produces an additional 31.09--62.61 points of degradation, again with every
provider-specific confidence interval below zero.
The frozen cohort contains no base supporting both stale and flipped faults, so
that comparison cannot be estimated on common support.

Thus, a larger \emph{proxy-score} stale effect persists relative to missing
and conflicting on the matched support. This does not establish a universal
fault-severity order; stale versus flipped is not identified.

\subsection{RQ3: Sensitivity Varies by Governance Event}

\begin{table*}[t]
\centering
\caption{
Base-equal, event-associated frozen proxy-score corruption/recovery
(in percentage points; corruption/recovery in each cell). Within each event,
we first average the supported fault effects for each base, then weight bases
equally. The three event columns therefore reproduce the primary event macro
in Table~\ref{tab:main-results} by an unweighted mean. Deletion coincides
with Memora; these are descriptive, not source-adjusted causal contrasts.
}
\label{tab:event-breakdown}
\small
\begin{tabular}{lccc}
\toprule
Model & Deletion (380 bases) & Temporal (169 bases) & Update (182 bases) \\
\midrule
DeepSeek V4 Pro & $-23.25/+22.63$ & $-24.65/+25.05$ & $-36.26/+36.63$ \\
GPT-5.6 Sol & $-13.68/+13.51$ & $-35.31/+35.50$ & $-44.32/+44.32$ \\
Claude Sonnet 5 & $-17.37/+17.02$ & $-26.53/+27.22$ & $-40.66/+40.84$ \\
Gemini 3 Flash & $-17.89/+17.72$ & $-37.08/+35.50$ & $-40.66/+40.66$ \\
Kimi K2.6 no-thinking & $-13.86/+16.23$ & $-23.96/+26.13$ & $-45.05/+43.41$ \\
Grok 4.3 no-reasoning & $-13.16/+11.67$ & $-19.92/+18.54$ & $-22.34/+22.53$ \\
\bottomrule
\end{tabular}
\end{table*}

The event means exactly reconstruct the main macro: for DeepSeek,
$(-23.25-24.65-36.26)/3=-28.05$ points. Its historical \emph{pair-weighted}
temporal estimate was $-28.93$, because 113 LongMemEval temporal bases
contribute two faults and 56 OAKS bases three. All-provider reconciliation
and original pair-weighted values remain in the supplemental audit.
Deletion equals Memora in this cohort and cannot identify an event effect
separately from source/task/scoring differences.

\subsection{RQ4: Oracle-Backed Enforcement Recovers Most of the Lost Performance}

Under the frozen proxy scorer, the controlled enforced condition offsets most
of the measured corruption-associated score drop for every provider.

The event-macro recovery effect ranges from
$+17.58$ points for Grok 4.3 to
$+31.29$ points for Gemini 3 Flash.
The descriptive six-provider mean is
\[
+27.51\ \text{percentage points}.
\]

Every provider-specific 95\% bootstrap confidence interval lies strictly above
zero.

For all six providers, enforced condition-level accuracy lies within
approximately 2.4 percentage points of the corresponding clean accuracy.
Thus, after benchmark-supported restoration, the deterministic proxy score
returns close to its clean baseline. The separate native-style Formal-300
assessment also shows recovery ($+32.67$ points, base-cluster 95\%
interval $[+26.42,+39.09]$) on its selected two-source panel. Neither estimate
establishes a population-level semantic recovery rate for the entire cohort.

These recovery effects measure the amount of degradation that is reversible
under oracle-backed control-state restoration.
They do not measure the ability of a deployment-time system to discover the
correct state autonomously.

\subsection{Control-Plane Propagation Diagnostic}

The offline propagation diagnostic shows that large frozen proxy-score
contrasts co-occur with changes to the operative model-visible projection.

Of the 2,080 supported base--fault pairs, 2,002
(96.25\%) have
\[
C^{\mathrm{corr}} \neq C^{\mathrm{clean}},
\]
while oracle-backed enforcement restores the clean operative projection in all
2,080 pairs (100\%).
These rates are computed over unique base--fault pairs;
the same deterministic projection is used for all six providers.

\begin{table*}[t]
\centering
\caption{
Offline propagation diagnostic by provider.
The two effect columns use frozen proxy-score corruption effects in percentage
points, stratified by whether the operative model-visible projection changes.
``Induced failure'' is the percentage of clean-correct, projection-changed cases
that become incorrect under corruption.
``Recovered'' is the percentage of those induced failures that return to
correct under oracle-backed enforcement.
}
\label{tab:propagation-diagnostic}
\small
\begin{tabular}{lrrrr}
\toprule
Model & Changed effect & Unchanged effect & Induced failure & Recovered \\
\midrule
DeepSeek V4 Pro       & -28.72 & -2.56 & 68.48\% & 97.05\% \\
GPT-5.6 Sol           & -27.57 &  0.00 & 69.58\% & 98.14\% \\
Claude Sonnet 5       & -26.67 & -5.13 & 71.58\% & 85.46\% \\
Gemini 3 Flash        & -29.32 & +1.28 & 73.43\% & 95.19\% \\
Kimi K2.6 no-thinking & -25.57 & -3.85 & 69.91\% & 89.07\% \\
Grok 4.3 no-reasoning & -18.38 & +2.56 & 76.90\% & 71.89\% \\
\bottomrule
\end{tabular}
\end{table*}

Across providers, the descriptive mean paired corruption effect is
$-26.04$ percentage points among pairs whose operative projection changes,
compared with only $-1.28$ points among pairs whose projection remains
unchanged.

The same pattern is visible when conditioning on cases that the model answers
correctly under clean governance.
Among clean-correct cases whose operative projection changes under corruption,
68.5\%--76.9\% become incorrect across the six providers.
Among these corruption-induced failures, oracle-backed enforcement restores
correctness in 71.9\%--98.1\% of cases.

The projection-change and projection-restoration counts are deterministic
control-plane observations; the induced-failure and recovery fractions use the
less-validated frozen answer proxy. Their conjunction is consistent with the
proposed pathway but does not independently establish semantic outcome recovery
or identify admission and context as separate mediators.

\subsection{Cross-Provider, Source, and Support Robustness}

The frozen proxy contrasts are negative under corruption and positive under
oracle-backed restoration for all six providers, despite different proxy
baselines. Stale has the largest unadjusted fault-associated proxy drop on
each provider's supported cases, while the smaller deletion-associated drop
is inseparable from the Memora source. Grok 4.3 has a lower proxy baseline and
smaller overall effect magnitude. These are cross-provider consistencies
\emph{conditional on the same scoring rule}, not independent semantic
replication or evidence of a source-adjusted severity ranking.

\begin{table*}[t]
\centering
\caption{
Source-level frozen proxy-score corruption/oracle recovery (points).
Each cell reports corruption/recovery.
Source estimates are descriptive because benchmark composition differs across
LongMemEval, Memora, and OAKS.
}
\label{tab:source-breakdown}
\small
\begin{tabular}{lccc}
\toprule
Model & LongMemEval & Memora & OAKS \\
\midrule
DeepSeek V4 Pro
& $-13.26/+11.96$ & $-23.25/+22.63$ & $-52.29/+54.38$ \\
GPT-5.6 Sol
& $-23.91/+23.91$ & $-13.68/+13.51$ & $-59.58/+59.79$ \\
Claude Sonnet 5
& $-21.09/+21.09$ & $-17.37/+17.02$ & $-50.63/+51.25$ \\
Gemini 3 Flash
& $-21.96/+20.87$ & $-17.89/+17.72$ & $-58.54/+58.33$ \\
Kimi K2.6 no-thinking
& $-16.30/+15.87$ & $-13.86/+16.23$ & $-58.75/+59.17$ \\
Grok 4.3 no-reasoning
& $+3.70/-4.35$ & $-13.16/+11.67$ & $-48.54/+48.13$ \\
\bottomrule
\end{tabular}
\end{table*}

Proxy effects remain negative for every provider on Memora and OAKS;
Grok's LongMemEval contrast instead reverses sign. OAKS has the largest
source-specific observed effect for every provider; document-clustered and
leave-one-document-out checks preserve that within-source direction. A
balanced set of 169 bases per event (507 total) retains the overall proxy
contrast directions, but it cannot decouple event type from benchmark source.

\subsection{Clean-Repeat Stochasticity Control}
\label{sec:clean-repeat}

To assess baseline repeat-generation variability without manipulating governance
state, we generated one additional response (Clean-B) for each of 60 frozen
clean inputs (Clean-A), holding the provider configuration and evaluation
criteria fixed. The paired sample contains 30 LongMemEval and 30 Memora blocks;
the former uses binary correctness, whereas the latter uses native continuous
FAMA. All 60 repeat responses were resolved with the frozen evaluator, using
88 planned judge requests, two protocol-triggered escalation judgments, and
one offline deterministic case.

\begin{table}[t]
\centering
\caption{Clean-repeat stochasticity control (60 paired blocks). $\Delta$ is Clean-A minus Clean-B, in percentage points. Confidence intervals use 10,000 paired-bootstrap resamples. LongMemEval uses binary correctness; Memora uses continuous native FAMA.}
\label{tab:clean-repeat}
\footnotesize
\setlength{\tabcolsep}{3.2pt}
\begin{tabular}{lrrrr}
\toprule
Subset & $n$ & Clean-A & Clean-B & $\Delta$ [95\% CI] \\
\midrule
Overall & 60 & .6906 & .6927 & $-0.21$ [$-11.16$, $+10.75$] \\
LongMemEval & 30 & .7333 & .7000 & $+3.33$ [$-16.67$, $+23.33$] \\
Memora & 30 & .6479 & .6853 & $-3.75$ [$-12.71$, $+6.25$] \\
\bottomrule
\end{tabular}
\end{table}

Within LongMemEval, 9/30 binary outcomes flip between clean generations
(70\% agreement), although their mean Clean-A minus Clean-B difference is
$+3.33$ points. Within Memora, 24/30 blocks share the same FAMA value and
the mean absolute FAMA change is $0.1041$; the mean signed difference is
$-3.75$ points. The pooled $-0.21$-point difference (95\% CI
$[-11.16,+10.75]$) mixes binary correctness and continuous FAMA, so we show
it only as a descriptive summary, not a common-unit variance estimate.

Formal-300 and clean-repeat use different cohorts and scoring units;
the control's wide intervals preclude equivalence or exact attribution claims.

\subsection{Governance-Evidence Audit}

The frozen-manifest consistency audit marks all 731 bases as supported:
191 LongMemEval, 380 Memora, and 160 OAKS-Novel. There are zero cases flagged
by its schema tests as unsupported, heuristic-only, ambiguous, or missing a
source locator. These \emph{rule-based audit outputs} reflect the frozen
evidence and support declarations used to build the intervention cohort.

\section{Discussion}

\paragraph{State integrity versus retrieval quality.}
A faithfully retrieved memory can be invalidated, superseded, or deleted.
Our interventions isolate governance projection changes while holding
stored evidence fixed. The selected LongMemEval/Memora native-style panel
supports damage and recovery \emph{within its respective outcome units};
OAKS option parity is a separate check, and full-panel effects remain
scorer-dependent proxies.

\paragraph{Confounding and heterogeneity.}
Stale-state corruption has the largest observed frozen-proxy damage on its
supported bases, including matched stale--missing and stale--conflicting
comparisons, but stale--flipped lacks common support. Deletion is present
only in Memora, precluding source-adjusted event rankings. Across the
six provider subsets, Formal-300 effects have consistent overall directions,
although small provider--source cells can have intervals spanning zero;
shared-base resampling does not make non-probability samples representative.
The full-panel provider agreement under the same proxy is therefore not six
independent confirmations of semantic accuracy. Conversely, the selected
native-style panel provides convergent evidence for the direction of damage
without calibrating the magnitude of the broader proxy results.

\paragraph{Recovery and architecture.}
Oracle-backed enforcement restores every audited model-visible projection,
but native-style Memora scoring retains a residual clean--enforced gap.
This is not evidence of autonomous inference of missing state or identification
of admission and context as distinct causal mediators. In practice,
memory systems should preserve provenance and current-version status, fail
closed on unverifiable control state, separate historical records from
operative records, and log admission decisions alongside model-visible
projections. The 60-block clean-repeat control shows that generation noise
is nonzero; its wide intervals do not support an exact decomposition of
intervention and stochastic effects. A deployment-ready extension would
need to verify the state reference without access to the benchmark oracle,
retain provenance for conflicting versions, and independently log whether
failure originated in state assignment, admission, context assembly, or the
model response. The present propagation audit covers the operative-context
boundary, not each stage as a separately estimated causal mediator.

\section{Limitations}

\paragraph{Scoring and sampling.}
The full-panel six-provider effects ($-27.55/+27.51$ points) use a frozen
Boolean correctness \emph{proxy}. Against independently evaluated native-style
answers, only 60/96 same-unit LongMemEval decisions agree
($\kappa=0.259$); Memora's 508 criteria spanning 42 answers cannot be
compared to Boolean proxy decisions as though they shared a unit. OAKS
native option parity holds on 6,714 resolved outputs, but does not validate
the mixed-source event macro. Formal-300 uses a deterministic,
non-probability selection (300 selected triplets; 298 resolved;
238 distinct bases) excluding OAKS. Its LongMemEval binary and
Memora continuous FAMA outcomes are not one shared accuracy metric; pooled
point differences and base-cluster bootstrap intervals are \emph{descriptive}
within this selected panel, not population precision. Small provider--source
cells, including DeepSeek--Memora and Grok--Memora, can have damage intervals
covering zero. The Memora enforced-minus-clean gap is $-4.35$ points.
No full-panel LongMemEval/Memora native re-evaluation was performed.

\paragraph{Identification and generality.}
Causal claims apply to paired governance interventions with task, evidence,
oracle, model, and protocol fixed, not to deployed fault prevalence or user
harm. Source/event/fault support is incomplete: deletion occurs only in
Memora, no base supports both stale and flipped, and source-specific validity
rules are not a universal authorization ontology. Balancing and matched
support cannot identify all independent severity effects. The offline
projection diagnostic does not observe admission and context as distinct
mediators. Providers also differ in API behavior and baseline performance.
Clean-repeat includes only one additional response per 60 blocks and has
wide intervals; its cross-source mixed-unit average is not a shared metric.

\paragraph{Audit and reproducibility boundaries.}
The de-identified case-trace example is a single existing frozen-proxy trace;
local v2--v3 correspondence was verified, but original text and model
responses cannot be publicly inspected. The 731/731 governance evidence
audit verifies provenance and schema consistency against the \emph{same}
construction assumptions, so shared interpretation errors remain possible.
The aggregate-only repository
ships code, prompt templates, configurations, hashes and summary statistics,
not licensed source texts, instantiated prompts, per-request outputs,
retry/timestamp histories, or adjudication records. It does not permit
independent end-to-end reproduction from the ZIP alone. Finally, oracle-backed
restoration assumes correct state is available and is not a deployment-ready
mechanism for authentication, malicious-state detection, or synchronization.

\section{Conclusion}

Accurate retrieval does not ensure that retained memory is valid to use.
Across 731 benchmark-grounded bases, state faults change the operative
projection in 96.25\% of supported pairs, and oracle-backed enforcement
restores the clean projection in 100\%. A separate native-style selected
panel resolves 298 LongMemEval/Memora triplets (238 distinct bases):
LongMemEval binary correctness shows $+40.54$ points of corruption-associated
damage and $+39.19$ points of recovery, while Memora native FAMA shows
$+30.60$ points of damage and $+26.24$ points of recovery on its rescaled
metric. All six generating-provider subsets have directionally positive
within-panel effects; base-cluster resampling preserves the qualitative
pattern, but some smaller subgroups remain uncertain and Memora shows a
residual recovery gap. OAKS native option parity holds on all 6,714 resolved
outputs. Neither the selected non-probability validation nor OAKS option
parity validates the full-panel frozen-scorer proxy contrasts
($-27.55/+27.51$ points) as native semantic population effects. These
bounded findings motivate explicit integrity checks for memory control
state alongside benchmark-appropriate downstream outcome evaluation.

\section*{Data and Code Availability}
An aggregate-only reproducibility package is available at
\url{https://github.com/chris3494285/govmem-paper-reproducibility}.
It provides code, prompt templates, tests, configurations, and summary
outputs. Licensed benchmark text, instantiated prompts, individual answers,
and private adjudication records are not redistributed; consequently, the
public release does not permit full case-level reruns. The repository
documents required upstream datasets, licensing restrictions, and
reproducibility boundaries.

\section*{Funding}
This research did not receive any specific grant from funding agencies in
the public, commercial, or not-for-profit sectors.

\section*{Declaration of generative AI and AI-assisted technologies in the manuscript preparation process}
During the preparation of this manuscript, OpenAI ChatGPT was used to assist
with manuscript organization, English-language editing, and \LaTeX{} drafting
and revision. The core research idea, experimental design, execution of
experiments, interpretation of results, and verification of the reported
findings were carried out by the authors. AI assistance did not constitute an
independent verification of experimental data or references. The authors
retain responsibility for reviewing the assisted material and for the
correctness of the submitted work.

\footnotesize
\setlength{\bibsep}{1pt}
\bibliographystyle{elsarticle-harv}
\bibliography{10_references}

\end{document}